\documentclass[conference]{IEEEtran}
\IEEEoverridecommandlockouts

\usepackage[english]{babel}
\usepackage[dvipsnames,svgnames,x11names]{xcolor}
\usepackage[a4paper, total={184mm,239mm}]{geometry}
\usepackage{amsmath,amssymb,amsfonts}
\usepackage{algorithmic}
\usepackage{graphicx}
\usepackage{textcomp}
\usepackage{csquotes}
\usepackage{booktabs}
\usepackage{xspace}
\usepackage{multirow}
\usepackage{subcaption}

\renewcommand{\baselinestretch}{0.99}  

\begin{document}
\bstctlcite{IEEEexample:BSTcontrol}

\newcommand{\backend}{backend\xspace}
\newcommand{\Dataset}{Dataset\xspace}
\newcommand{\dataset}{dataset\xspace}
\newcommand{\opencl}{OpenCL\xspace}
\newcommand{\openmp}{OpenMP\xspace}
\newcommand{\pulp}{PULP\xspace}
\newcommand{\riscy}{RI5CY\xspace}
\newcommand{\runtime}{runtime\xspace}

\title{Source Code Classification for Energy Efficiency in Parallel Ultra Low-Power Microcontrollers}

\author{%
\IEEEauthorblockN{Emanuele Parisi, Francesco Barchi, Andrea Bartolini, Giuseppe Tagliavini, Andrea Acquaviva}
\IEEEauthorblockA{\textit{DEI}
\textit{Università di Bologna}
, Bologna, Italy \\
emanuele.parisi@unibo.it}
}

\maketitle

\begin{abstract}
The analysis of source code through machine learning techniques is an increasingly explored research topic aiming at increasing smartness in the software toolchain to exploit modern architectures in the best possible way. In the case of low-power, parallel embedded architectures, this means finding the configuration, for instance in terms of the number of cores, leading to minimum energy consumption. Depending on the kernel to be executed, the energy optimal scaling configuration is not trivial. 
While recent work has focused on general-purpose systems to learn and predict the best execution target in terms of the execution time of a snippet of code or kernel (e.g. offload \opencl kernel on multicore CPU or GPU), in this work we focus on static compile-time features to assess if they can be successfully used to predict the minimum energy configuration on PULP, an ultra-low-power architecture featuring an on-chip cluster of RISC-V processors. 
Experiments show that using machine learning models on the source code to select the best energy scaling configuration automatically is viable and has the potential to be used in the context of automatic system configuration for energy minimisation.
\end{abstract}


\section{Introduction}
\label{section:introduction}

Understanding the impact of a source code fragment on a given target architecture is an interesting problem considering the increasing complexity and parallelism of embedded platforms, as it opens the way to automatic configuration and optimisation strategies. 

Considering ultra-low-power parallel architectures targeting 1 GOPS/mW performance/power envelope, they leverage scalable parallelism, optimised memory access patterns and flexible low power states such as clock and power gating~\cite{pulp15} to reach their efficiency target. Depending on the computation to be executed, the minimum energy configuration in terms of the number of cores depends on the pressure imposed on the processing and memory components and their power consumption in the various functional states. Also, the best energy configuration is usually different from the one leading to higher speed-up. 

In this work, we study how the optimal trade-off can be derived directly from source code analysis at compile time and the informative gap with profiling information that can be obtained by profiling the execution on the target platform.

Approaches based on source code analysis have been applied to take decisions on parallelism mapping~\cite{wang14}, thread coarsening~\cite{magni14}, or offloading decisions on general-purpose GPU based systems \cite{grewe2013,cummins2017,barchi2019}.
However, these techniques have never focused on energy, nor they targeted ultra-low-power embedded architectures.
From the other side, previous research work investigated power and energy modelling of parallel architectures using features extracted from code execution profiling~\cite{shajulin15,rejitha17}.

In the present work, the target is to predict the scaling configuration (i.e. the number of parallel running cores) providing minimum energy consumption using source code information only. We modelled this as a classification task.
The purpose of the classifier is to assign each computational kernel to its minimum energy class. 

To achieve this target, we built a dataset composed by standard \openmp benchmarks augmented with a set of custom parametric kernels that we designed to stimulate the energy trade-offs of the target architecture.
The dataset has been used to train a decision tree model. 
The benchmarks have been ported to the PULP platform~\cite{pulp15}, the open-source ultra-low-power parallel architecture we considered in this work. 
The results obtained for the PULP platform are general for the same class of devices: which leverages parallelism and low-power design for energy-efficiency.

We first defined a set of features that we extracted by parsing the LLVM-Intermediate Representation and tailored to assess the energy trade-off on the target architecture. Then we exploited additional features obtained from an existing LLVM code analysis tool.

Also, we compared the classification accuracy obtained by using compile-time (i.e. static) features extracted from the source code with respect to profile-based (i.e. dynamic) ones. 
This comparison is crucial as dynamic profiling information such as memory contention and low-power states transitions may be very relevant for the energy trade-off. 
Results show that using only static features can reach more than 85\% energy classification accuracy when tolerating 8\% of the energy impact of miss-classification. 
Our experiments show that the accuracy gap between static and dynamic features is lower than 10\% in the present dataset.

The contribution of the work can be summarised as follows: i) We designed a dataset of kernels for source code energy classification in parallel architectures; ii) We stated that the energy classification problem is not a trivial extension of performance or speed-up classification; iii) We demonstrated that source code energy classification is feasible, and we quantified the accuracy gap with classification based on dynamic features.

The rest of the paper is organised as follows: Section \ref{section:background} describes PULP and reviews source code and energy models. 
In section \ref{section:methods}, we describe source code analysis method, while in section \ref{section:results} we discuss the obtained results.

\section{Background and Related Work}
\label{section:background}

\subsection{The Parallel Ultra-Low-Power Platform}
The target architecture of this work is the Parallel Ultra-Low-Power Platform (\pulp), a soft IP implementing a cluster of processors built around a parametric number of \riscy \cite{gautschi2017near} cores (up to 16).
\riscy is a RISC-V based processor with dedicated extensions for Digital Signal Processing (DSP) and machine learning workloads.
The cores share a multi-banked scratchpad memory called Tightly-Coupled Data Memory (TCDM), enabling single-cycle data access and allowing data-parallel programming models such as \openmp.
Outside the cluster, the architecture features an L2 memory hierarchy level, composed of a 15-cycle latency multi-banked scratchpad memory. 
A DMA enables data transfers between the two memory levels.

We consider a PULP instance including 8 cores, a 512 KiB L2 memory and a 64 KiB TCDM.
The cluster cores are connected to 4 Floating Point Units (FPUs), which are shared among the cores in the cluster using an interconnect that enables a fixed mapping of cores to available FPUs.
The FPU architecture is pipelined with a single stage.
This architecture, introduced in \cite{montagna2020transprecision} as \emph{8c4f1p} (8 cores, 4 floating-point units, 1 pipeline stage), is the most energy-efficient configuration of PULP; experimental results show that this solution outperforms its main competitors in the domain of embedded processing systems.

\subsection{Machine learning for source code performance estimation}

The topic of machine learning based source code analysis is gaining increasing interest in recent years.
The outbreak of complex heterogeneous architectures inspired many works that aim at exploiting source code features for predicting the device with the shortest \runtime where to execute a computation.
%

Authors of \cite{grewe2013} predict whether a given \opencl kernel is most suited for running on CPU or GPU. 
They exploit a decision tree, a standard machine learning technique that supports decisions by checking a sequence of control statements. 
That work shows promising classification accuracy considering static source code features such as opcode counts, kind of memory accesses, and amount of processed data.

Successively, \cite{cummins2017, barchi2019, brauckmann2020, barchi2020exploration} improves accuracy results obtained by \cite{grewe2013} exploiting deep learning models based on LSTM (Long Short-Term Memory). 
Even if promising, deep learning models do not allow insight into what static source code features are most significant for carrying out the classification task.

\subsection{Energy estimation from source code} 

A number of literature papers provide methodologies for inferring the energy consumption of an application by looking at its source code, avoiding complex and time-consuming RTL simulations or measurement campaigns.
While it is relatively common in High-Performance Computing (HPC) and embedded High-Performance Computing (eHPC) systems to have a power gauge, this is not the case of embedded systems where power consumption can only be accessed in a lab setting.

Such work can be divided into two families depending on how do they approach source code: methods based on dynamic features and methods that exploit static analysis.

In general, collecting dynamic features requires to run the program for accessing performance counters or the real trace of opcodes executed.
Dynamic features tend to be more accurate, but it is not always possible to collect them.  As an example, authors in \cite{tiwari1994power, kerrison2015energy} assign an average current/power cost to every opcode in the target ISA and applies it to the execution trace of the program to estimate the energy consumption. 
The authors of \cite{shajulin15} combine performance counters values (monitored during the application run) and a random forests model for predicting the energy consumed by parallel \openmp applications.

Differently, static analysis exploits metrics available without running the program, such as data-flow analysis or opcode family counts in a section of source code.
The authors of \cite{grech2015static} present a static analysis method to predict energy consumption based on the data-flow analysis.
It extracts and solves \textit{cost-relations} from the code and expresses the energy required to execute the kernel as a function of the amount of data to be processed.
Unfortunately, a generalisation of such methodology in multi-core environments is not available.

To the best of our knowledge, no static methods provide prediction models for detecting the optimal parallelism in \openmp applications using only static source code features.
Here, we take into account low-power multi-core environments and provide a detailed energy estimation that considers contention on shared resources and advanced core power management policies such as clock gating.
Moreover, we demonstrate that it is possible to exploit machine code analyser tools such as LLVM-MCA in order to infer information about energy and parallelism efficiency from source code behaviour in processor microarchitectures.

\section{Methods}
\label{section:methods}
\begin{figure*}[t]
    \centering
    \includegraphics[width=6.5in]{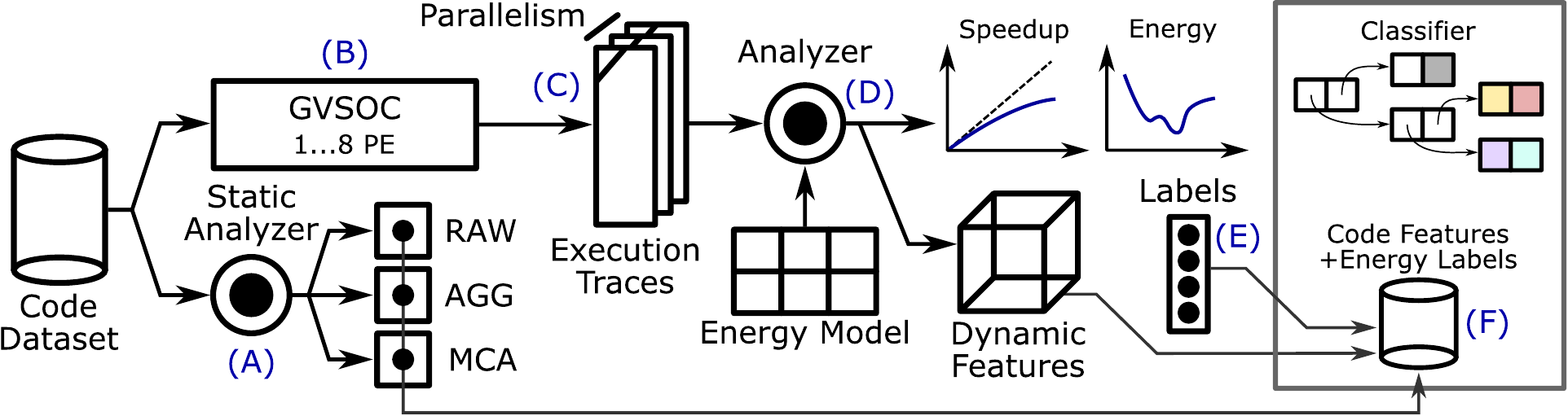}
    \caption{
        Workflow to identify the minimum energy parallelism on a PULP cluster and to define a dataset composed of static and dynamic features. 
    }
    \label{figure:pulp_energy_flow}
    \vspace{-0.20in}
\end{figure*}

\subsection{Methodology}
\label{subsection:methodology}

In embedded parallel processors, software energy efficiency is achieved exploiting hardware parallelism. With the increase of the number of cores used by the application the \runtime decreases (as well as the leakage energy) whereas the dynamic power increases. 

In this paper, we aim at proving that a machine learning model fed with static source code information is able to learn the best configuration for optimal energy consumption on parallel embedded microcontrollers.
Figure \ref{figure:pulp_energy_flow} details the proposed approach.
It consists of the following steps:

\textit{(A)} A preliminary features extraction activity is performed on all samples in the dataset through static source code analysis.
    Details about the dataset construction and the machine learning features are provided in Sections \ref{subsection:dataset_description} and \ref{subsection:features_selection}.

\textit{(B)} Each sample in the dataset is analysed using a cycle-accurate PULP simulator that provides execution traces for keeping track of opcodes executed, memory transactions, active wait cycles, and cores idleness due to clock gating.

\textit{(C)} All samples are simulated eight times using an increasing number of the cores available in PULP.

\textit{(D)} All execution traces are combined with the energy model of Table \ref{table:energy_model}.
    This allows assigning an energy cost to the execution of each sample as a function of the number of used cores.
    The energy model is described in Section \ref{subsection:energy_model}.

\textit{(E)} The number of threads to be used for minimising energy consumption is used for labelling each sample.

\textit{(F)} The collection of labelled samples, each with its static features, represents the dataset for training the decision tree.

\subsection{Dataset description}
\label{subsection:dataset_description}
A collection of parallel programs to be measured and analysed has been defined.
We choose to express kernel parallelism with \openmp, a widely-used programming model for shared-memory architectures supported by an increasing number of platforms, including PULP.

Considering the \openmp standard, it is common for embedded research-oriented architectures not to implement the full programming model standard, rather a subset of functionalities for supporting the most common scenarios \cite{barchi2017efficient}. 
For this reason, in this work, we had to customise application kernels constituting the dataset carefully and, in most cases, discard publicly available \openmp datasets. 

In the case of PULP, the current OpenMP runtime does not implement tasking and supports a limited subset of loop scheduling policies.
Concerning memory allocation, PULP provides a comprehensive but non-standard set of interfaces for enabling on-cluster data allocation.
It allows taking advantage of fast access in TCDM without the burden of explicitly programming DMA transfers from the off-cluster memory, which is the default target for dynamic memory allocation.
We also transformed the benchmarks to make them parametric concerning the type of data manipulated during computation.
In fact, embedded systems may have constrained hardware resources that may cause the same program to behave very differently depending on the kind of data it has to deal with.
This is especially true when dealing with floating-point operations, which are often performed on a resource contended by cores (FPU).

\subsection{Energy model}
\label{subsection:energy_model}

The energy model we used is detailed in Table \ref{table:energy_model}.
The energy contribution of each component of the PULP cluster is characterised in terms of both leakage and switching activity.
The energy consumption due to processing elements depends on the classes of opcodes executed and on the number of active wait cycles (NOP) executed. 
Additionally, also advanced low-power states are considered when the core is driven in clock-gating to reduce power consumption during periods of inactivity.
Memory, FPU, and DMA models distinguish between active and idle power consumption.
Moreover, memory models differentiate the energy cost paid due to read and write operations.
Additionally, further costs are considered for taking into account energy consumption due to not explicitly modelled circuitry within the PULP cluster, such as the cores-to-TCDM bus and the event unit, which manages power gating and interrupts dispatching.

\subsection{Feature selection}
\label{subsection:features_selection}

\begin{table}[]
\caption{PULP Energy model}
\label{table:energy_model}
\centering
\begin{tabular}{@{}lr@{}}
 \toprule
 Operating Region & Energy [fJ]\\
 \toprule
 \multicolumn{2}{@{}c@{}}{Processing Element} \\
 \midrule
Leakage & 182 \\
 NOP &  1212 \\
 ALU &  2558 \\
 FP & 2468 \\
 L1 & 3242 \\
 L2 & 1011 \\
 CG &  20 \\
 \toprule
 \multicolumn{2}{@{}c@{}}{FPU} \\
 \midrule
 Leakage & 191\\
 Operative & 299\\
 Idle & 0 \\ 
 \toprule
 \multicolumn{2}{@{}c@{}}{Other Cluster Components} \\
 \midrule
 Leakage & 655 \\
 Active &  2702 \\
 \bottomrule
 & \\
 & \\
 & \\
 & \\
 \end{tabular}
\hspace{0.02in}
\begin{tabular}{@{}lr@{}}
 \toprule
 Operating Region & Energy [fJ]\\
 \toprule
 \multicolumn{2}{@{}c@{}}{Memory Bank L1} \\
 \midrule
 Leakage & 49\\
 Read & 2543\\
 Write & 2568\\ 
 Idle & 64\\
 \toprule
 \multicolumn{2}{@{}c@{}}{Memory Bank L2} \\
 \midrule
 Leakage & 105\\
 Read & 2942\\
 Write & 3480\\ 
 Idle & 13\\
 \toprule
 \multicolumn{2}{@{}c@{}}{ICache} \\
 \midrule
 Leakage & 774\\
 Use & 4492\\
 Refill & 5932\\ 
 \toprule
 \multicolumn{2}{@{}c@{}}{DMA} \\
 \midrule
 Leakage & 165\\
 Transfer & 1750\\
 Idle & 46\\ 
 \bottomrule
\end{tabular}
\vspace{-0.15in}
\end{table}

In this section, we describe the features we considered for training the classification model, which is based on a decision tree. 
We considered two ensembles of static source code features: The ones introduced by the authors of \cite{grewe2013} and the ones provided by machine code analyser tools such as LLVM-MCA.
Both families of metrics are summarised in Table \ref{table:static_features_selection}, labelled as RAW, AGG (aggregate) and MCA (machine code analyser).

The authors of \cite{grewe2013} considered a set of six RAW metrics for the static analysis of \opencl kernels. 
Such metrics were then combined into the four features that are actually used to feed the decision tree.
However, in the context of deeply embedded systems, not all the RAW metrics defined in \cite{grewe2013} can be used.
First of all, we do not consider the distinction between global and local memory accesses, assuming that all data are in TCDM.
Such an assumption is reasonable since an architecture like PULP works at its maximum efficiency if data accesses happen in the on-cluster TCDM.
Part of the complexity of dealing with embedded devices of the same class of PULP is to carefully program DMA transfers from off-cluster memory such that transfers and computation are overlapped in time.
Moreover, considering coalescing is not useful in our case since scratchpad memories are not sensible to access patterns. 
Finally, the average number of work-items of a kernel is a metric specific to the \opencl programming model. 
This does not apply to \openmp codes.
Here we do propose to consider instead the average number of iterations that can be carried concurrently in \openmp parallel regions within the kernel.
For combining the RAW metrics into the four AGGregate static features, we remain consistent with what is described in \cite{grewe2013} and summarised in Table \ref{table:static_features_selection}.

\begin{table}[]
\captionsetup[subtable]{margin=0pt,justification=raggedright,position=top}

\caption{Static Features} \label{table:static_features_selection}

\begin{subtable}{1.00\linewidth}{
 \caption{RAW and AGG features} \label{table:static_features_selection}
 \begin{tabular}{@{}lll@{}}
 \toprule
 \multicolumn{2}{@{}c}{Features} &  \multicolumn{1}{l@{}}{\multirow{2}{*}{Notes}} \\
 \cmidrule(lr){1-2}
 \multicolumn{1}{@{}l}{\cite{grewe2013}} & \multicolumn{1}{l}{This work} &  \\
 \toprule
 \multicolumn{3}{@{}c@{}}{\textbf{RAW}} \\ 
 \midrule
 \texttt{comp} &
 \texttt{op} & 
 \# of ALU, FP and JUMP opcodes \\
 \texttt{mem} &
 \texttt{-} & 
 Not used \\
 \texttt{localmem} &
 \texttt{tcdm} & 
 \# of accesses in on-cluster TCDM memory \\
 \texttt{coalesced} &
 \texttt{-} & 
 Not meaningful on the PULP architecture \\
 \texttt{transfer} &
 \texttt{transfer} & 
 Amount of data the kernel works on \\
 \texttt{avgws} &
 \texttt{avgws} & 
 Average \# of iterations in parallel regions \\
 \toprule
 \multicolumn{3}{@{}c@{}}{\textbf{AGG}} \\ 
 \midrule
 \texttt{F1} &
 \texttt{F1} &
 \texttt{transfer} / (\texttt{op} + \texttt{tcdm}) \\
 \texttt{F2} &
 \texttt{-} &
 Not available, depends on \texttt{coalesced} \\
 \texttt{F3} &
 \texttt{F3} &
 \texttt{avgws} \\
 \texttt{F4} &
 \texttt{F4} &
 \texttt{op} / \texttt{tcdm} \\
 \bottomrule
\end{tabular}}
\end{subtable}

\vspace{0.1in}
\begin{subtable}{1.0\linewidth}{
\caption{MCA features} \label{table:static_features_selection}
\begin{tabular}{@{}ll@{}}
  \toprule
  Features & Notes \\
  \toprule
  \multicolumn{2}{@{}c@{}}{\textbf{MCA}} \\ 
  \midrule
  \texttt{uOPSpc} & Micro operations issued per cycle \\
  \texttt{IPC} & Instructions per cycle \\
  \texttt{RBP} & Reverse block throughput \\
  \texttt{RPDiv} & Resource pressure on the divider port \\
  \texttt{RPFPDiv} & Resource pressure on the floating-point divider port \\
  \texttt{RP0} & Resource pressure on Port\,0 (Other Components) \\
  \texttt{RP1} & Resource pressure on Port\,1 (Other Components) \\
  \texttt{RP2} & Resource pressure on Port\,2 (AGU, Load Data) \\
  \texttt{RP3} & Resource pressure on Port\,3 (AGU, Load Data) \\
  \texttt{RP4} & Resource pressure on Port\,4 (Store Data) \\
  \texttt{RP5} & Resource pressure on Port\,5  (INT-ALU, INT Vec. ALU, LEA) \\
  \texttt{RP6} & Resource pressure on Port-6 (INT-ALU, Branch) \\
  \texttt{RP7} & Resource pressure on Port-7 (Address Generation Unit) \\
 \bottomrule
\end{tabular}}
\end{subtable}
\end{table}

The LLVM framework provides a tool for static machine code analysis called LLVM-MCA. 
It models the execution engine of many out-of-order microarchitectures and provides insights about how a set of opcodes are dispatched to the various execution units, or \textit{ports}, assuming cache hits and perfect branch predictions. 
As a result, LLVM-MCA provides a set of metrics, called \textit{port pressures}, which describe how much the analysed flow of instructions stimulates the execution units.
Given the ease of collecting such features, readily available within the LLVM framework, we test whether they can be used as static kernel \textit{fingerprint} able to help the decision tree classifier in modelling the source code of the kernel to be analysed for solving our domain-specific problem.

\begin{table}[]
\caption{Dynamic Features}
\label{table:dynamic_features_selection}
\centering
\begin{tabular}{@{}lp{2.5in}@{}}
 \toprule
  \multicolumn{1}{@{}l}{Features} & \multicolumn{1}{l@{}}{Notes} \\
 \midrule
 \texttt{PE\_idle} & Fraction of cycles in which a core incurs in resource contention or in a multi-cycle instruction. \\
 \texttt{PE\_sleep} & Fraction of cycles in which a core is in clock-gating. \\
 \texttt{PE\_alu}  & \# of opcodes involving the usage of the ALU. \\
 \texttt{PE\_fp}  & \# of opcodes involving the use of the FPU. \\
 \texttt{PE\_l1}   & \# of opcodes involving access to the TCDM. The access level is inferred intercepting the address required by the operation at runtime. \\
 \texttt{PE\_l2}   & \# of opcodes involving an access to off-cluster memory. \\
 \midrule
 \texttt{L1\_idle} & \# of cycles in which a TCDM bank is idle. \\
 \texttt{L1\_read} & \# of read request received by a TCDM bank. \\
 \texttt{L1\_write} & \# of write request received by a TCDM bank. \\
 \texttt{L1\_conflicts} & \# of contemporary requests received by a TCDM bank. \\
 \bottomrule
 \end{tabular}
\end{table}

\section{Results}
\label{section:results}

\subsection{Test Bed}
GVSOC is the virtual platform included in the PULP-SDK.
It is fast compared to an RTL simulation and provides a good cycle accuracy.
Such properties are key requirements for integration into development flows.
The virtual platform also provides execution traces that describe the status of cluster components during the program execution.

The power numbers have been derived by a post place-and-route analysis with Synopsys PrimeTime 2019.12, setting a nominal voltage of 0.65 V and extracting value change dump (VCD) traces through parasitic-annotated post-layout simulation of synthetic benchmarks using Mentor Modelsim 2008.06.
These numbers include components for static and dynamic power consumption. 
Since each synthetic benchmark includes a single class of instructions, these values can be integrated to provide the energy consumption associated with a specific class.

We used GVSOC to get the execution traces and infer the energy consumed when running an \openmp kernel on PULP.
The traces are a dump of events triggered by the components modelled by the virtual platform.
Each component is identified by a path that indicates its position within the architecture.
The trace analysis software aims to identify all the events related to the useful components for energy calculation.
It consists of two modules, a hierarchical set of listeners and a trace-analyser.
The listeners are aggregated within the PULPListeners class, which exposes methods to query the status of the platform and its components.
PULPListeners contains 8 CoreListeners, 16 L1BankListeners and 32 L2BankListeners.
Each listener registers itself on the trace-analyser providing the path needed to capture the events intended for it.

The trace-analyser reads the GVSOC trace line by line and parses it using regular expressions to obtain: the event cycle number, the path of the component that issued the event, and other information that will be analysed later by a listener.
CoreListeners get events from ``cluster/pe/insn'' to analyse the opcodes trace and on ``cluster/pe/trace'' to identify clock gating regions and wait cycles.
The BankListeners get events from ``cluster/l1/bank/trace'' to analyse writing and reading events on the bank and the number of conflicts that occur whenever multiple requests are received in the same cycle.

After analysing the trace, it is possible to filter out events within a range of cycles.
The procedure involves identifying the range of cycles in which the parallel code fragment is contained (function ``void kernel(...)'').
Within the region, the dynamic features listed in Table \ref{table:dynamic_features_selection} are identified, and the energy contributions associated with each operating state of a component is counted as described in Table \ref{table:energy_model}.

\subsection{Dataset analysis}

The \openmp dataset we use consists of a collection of three suites of benchmarks, for a total of 59 distinct kernels written in C. 
The suites of benchmarks for the sake of our analysis are \textit{Polybench}, \textit{UTDSP}, and \textit{Custom}. 
\textit{Polybench} is a well-known set of programs for testing polyhedral optimisation passes in compilers. 
\textit{UTDSP} comprises a set of kernels designed for testing optimisation targeting digital signal processors. 
At last, we added in our \dataset a collection of hand-written kernels designed to stimulate different patterns of memory accesses, compute operations, and synchronisation primitives.

Each kernel is parametric concerning the type of data it deals with and the amount of data it processes. 
Concerning data types, we considered 32-bits integers and 32-bits single-precision floating points. 
We avoid using double precision floating-points since the processing elements inside PULP does not support them.
Moreover, we leave the impact of compact integer types, such as 16 or 8 bits integers, for later works.

The execution of each kernel, instantiated with a specific type, is repeated multiple times with the different amount of processing data, for checking how problem size impacts energy efficiency. 
For each kernel, we tested a problem size of 512, 2048, 8196, and 32768 bytes. 
The quantity and variety of chosen payload size have two advantages. 
On the one hand, it reflects a typical payload size suitable for the amount of computation in a parallel microcontroller of the power class of PULP.
On the other hand, such a choice allows us to fit all the data the benchmarks work on in the scratchpad memory.
In this way, we avoid the need to take into account DMA transfers from the off-cluster memory to the scratchpad, which would make the energy analysis notably more difficult.
Under the assumptions above, the dataset of kernels we used to train and test the machine learning model is composed of 448 samples. 
The dataset shows a class unbalance between 5\% and 15\%, except for the class with label ``8'' which accounts for the 34.8\% of the samples collection.

When evaluating the performance of the classifier, we considered classification accuracy as metrics.
However, we also considered that in some cases, selecting a number of processing elements that leads to a small amount of energy wasted with respect to the theoretical minimum may be acceptable from the engineering point of view. 
We computed the accuracy with and increasing tolerance threshold on the energy wasted in case of misclassification. 
For example, imagine we are interested in evaluating the performance of the classifier with an energy tolerance threshold equal to $t$. 
If a dataset sample is the most energy-efficient when parallelised with four processing elements, but the classifier predicts that it should be computed with six processing elements, such a prediction is considered correct if the energy wasted running that kernel with six cores instead of 4 is lower than $t\%$.

Every training experiment described in the following section is performed with 10-fold stratified cross-validation.
Moreover, each cross-validation was repeated 100 times with random seeds, for ensuring to get unbiased accuracy results.

\subsection{Optimal configuration selection}
\begin{figure}[t]
    \includegraphics[width=3.5in]{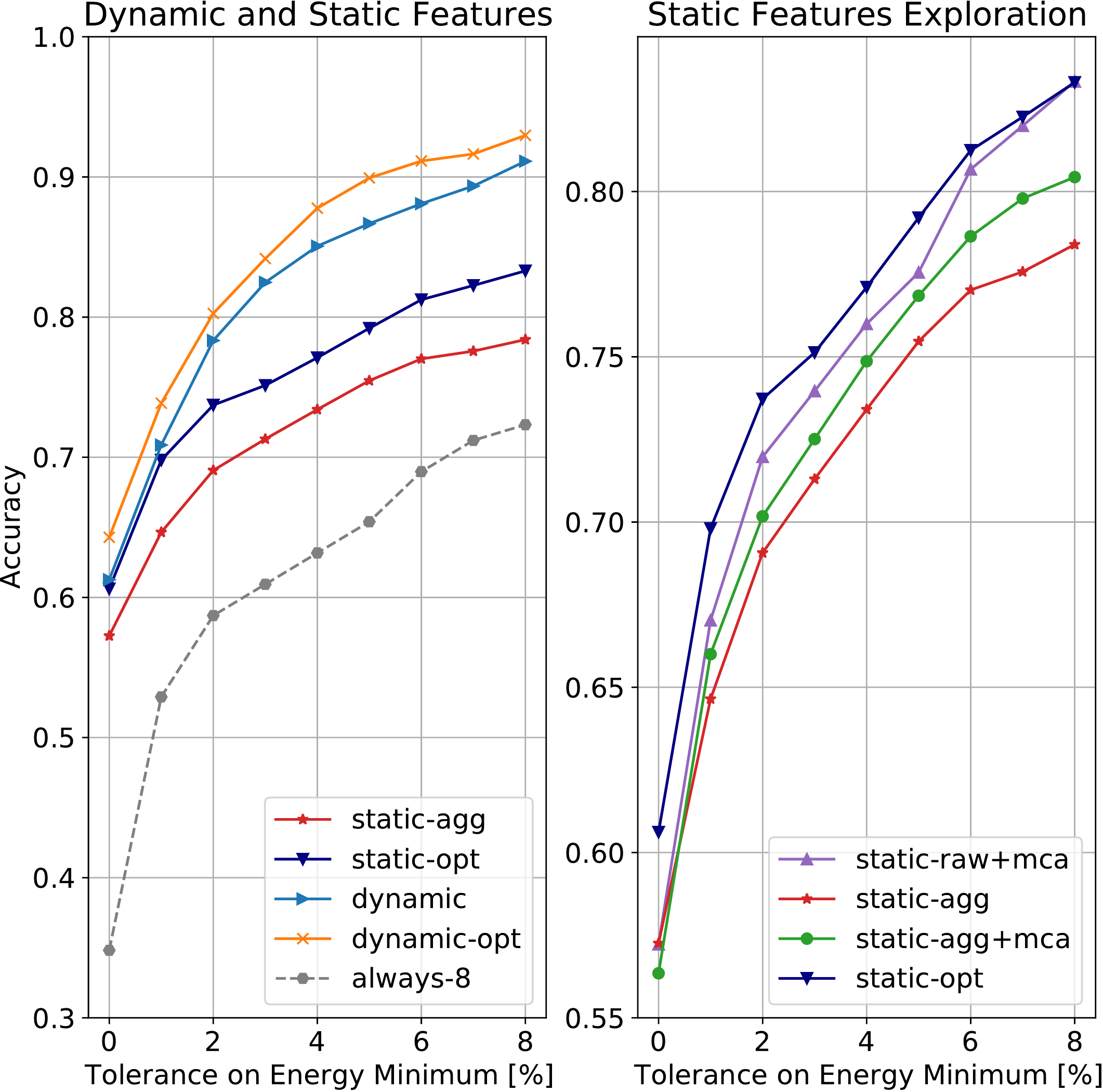}
    \caption{
        The image represents the classification accuracy obtained by a Decision Tree when the percentage tolerance on the energy minimum varies.
        The first graph shows static and dynamic features against ``always-8'' choice.
        The second graph shows the classification accuracy on different static features.
    }
    \label{figure:classification_results}
\end{figure}

In our work, we test the capability of machine learning models to infer the most energy-efficient parallelism configuration on PULP.
Our investigation is about if it is possible to feed the classifier only with features extracted by a static code analysis.
Such exploration is carried out in three distinct steps: i) Preliminary analysis of aggregate (AGG) static features ii) Analysis exploiting dynamic features coming from GVSOC traces and selection of most promising classification features iii) Optimisation of static features.

First of all, we check the classification accuracy of the decision tree fed with static features as detailed in Section \ref{section:methods}. 

At first, we consider the aggregate (AGG) set of features (F1, F3 and F4), described in Table \ref{table:static_features_selection}. 
We compare our result with a naive classifier using all the processing elements in the cluster (always-8).
The comparison can be appreciated in the left plot of Figure \ref{figure:classification_results}, which stresses that the red line always outperforms the dashed grey line.
Specifically, considering a tolerance of 5\% on energy wasted leads to a classification accuracy higher than 75\%.
The exploration of dynamic features is crucial to identify new static features necessary to improve classification performance.

Then we perform the same experiments using dynamic features extracted from GVSOC. 
Since traces are used to compute the energy consumption of a program, they contain the "ground truth" to identify the best energy parallelism. 
Since we expect dynamic features to be more informative than static ones, we want to identify the optimal subset, which enables better classification results.
Dynamic features are sorted according to the importance the decision tree assigns them.
From the analysis, a set of important features are listed in Table \ref{table:dynamic_features_importance}.

\begin{table}[]
\caption{Most Relevant Features}
\label{table:dynamic_features_importance}
\centering
\begin{tabular}{@{}lrr|lrr@{}}
 \toprule
 Label & PEs & Importance & Label & PEs & Importance \\
 \midrule
 \multicolumn{6}{@{}c@{}}{Dynamic Features}  \\ 
 \midrule
 PE\_sleep     & 8 & 19.6 \% & PE\_sleep     & 5 &  3.5 \% \\
 PE\_sleep     & 2 & 11.7 \% & L1\_conflicts & 5 &  3.2 \% \\
 PE\_idle      & 5 &  6.8 \% & PE\_sleep     & 6 &  3.1 \% \\
 L1\_write     & 1 &  6.7 \% & PE\_alu       & 6 &  2.3 \% \\
 L1\_conflicts & 6 &  4.1 \% & PE\_sleep     & 7 &  2.1 \% \\
 L1\_read      & 8 &  4.0 \% & PE\_idle      & 3 &  1.9 \% \\
 \midrule
 \multicolumn{6}{@{}c@{}}{Static Features}  \\ 
 \midrule
 avgws &  & 19.6 \% & RP-4   &  & 3.5 \% \\
 F4    &  & 11.7 \% & uOPSpc &  & 3.2 \% \\
 F1    &  &  6.8 \% & RP-7   &  & 3.1 \% \\
 \bottomrule
 \end{tabular}
\end{table}

The most relevant is the PE\_sleep feature, which represents the clock-gating cycles computed with a parallelism of 8 and 2 cores.
Those two values are important since they discriminate the source code behaviour with minimum and maximum parallelism.
Other relevant features are PE\_idle using five cores and L1\_write operations without parallelism, which respectively identify the wait cycles using half of the available parallelism and the number of memory writes without parallelism.

The left plot of Figure \ref{figure:classification_results} highlights the classification accuracy over 8 classes using different combinations of the static features detailed in \ref{section:methods}. 
The accuracy with an energy tolerance threshold of 0\% is substantially coherent and approximately equal to 57\%. 
Interestingly, allowing an energy threshold tolerance of 5\%, which is feasible in most cases, the classification accuracy approaches 80\%. 
Scoring the features used by the decision tree by importance and pruning less informative ones allows getting an "optimised" classifier that reaches 61\% accuracy without a threshold and 79\% with a 5\% threshold over eight classes.

\section{Conclusions}
\label{section:conclusions}

Automatic source code configuration is a problem that gains interest as architectures become more and more complex and heterogeneous.
This work represents the first attempt to predict the optimal number of cores for minimising the execution energy of \openmp kernels on deeply embedded architectures using static code analysis. 
We targeted the PULP architecture, a state-of-art parallel ultra-low-power embedded microcontroller. 
We feed the decision tree model with dynamic features to highlight the most promising ones for making the static features classifier more robust. 
Finally, we show that a decision tree fed with static source code features reaches a substantial accuracy of 61\%. 
Accuracy approaches 80\% if a 5\% tolerance threshold on energy wasted is introduced when evaluating the classifier.

We plan to extend this work improving the dataset coverage; increasing the number of kernels and considering different parallel programming models.
Moreover, we will model DMA transfers and memory hierarchy, and we will leverage deep learning models able to enhance the prediction capabilities offered by the solutions proposed by the current work.

\section*{Acknowledgment}
This work was supported in part by the Italian Ministry for Education, University and Research (MIUR) under the program “Dipartimenti di Eccellenza” (2018–2022).

\tiny
\bibliographystyle{IEEEtran}
\bibliography{bibliography}

\end{document}